\documentclass[12pt,conference]{IEEEtran}
\ifCLASSOPTIONcompsoc
\else
\fi
\ifCLASSINFOpdf
  \usepackage[pdftex]{graphicx}
\else
\fi
\usepackage[usenames, dvipsnames]{color}
\usepackage{amsmath}
\usepackage{amssymb}

\usepackage{caption}
\usepackage{balance}
\usepackage{subcaption}
\usepackage{listings}
\usepackage{fancyheadings,color}
\usepackage[font=small,labelfont=bf]{caption}
\makeatletter
\def\@seccntformatinl#1{\csname the#1dis\endcsname\hskip 1em\relax}
\makeatother
\lstdefinestyle{mystyle}{
    basicstyle=\footnotesize,
    breakatwhitespace=false,         
    breaklines=true,                 
    captionpos=b,                    
    keepspaces=true,                 
    numbersep=5pt,                  
    showspaces=false,                
    showstringspaces=false,
    showtabs=false,                  
    tabsize=2
}
\begin{document}

\pagestyle{fancy}
\chead{\footnotesize \textcolor{blue}{To be published at the 2018 IEEE Symp. Series on Comp. Intelligence (IEEE SSCI 2018), 18-21 NOV, 2018, BENGALURU, INDIA}}
\rhead{}
\lhead{}
 \renewcommand{\headrulewidth}{1pt}
\setlength{\headheight}{10pt} 

\title{Facial Recognition with\\ Encoded Local Projections\\
}

\author{\IEEEauthorblockN{Dhruv Sharma, Sarim Zafar}
\IEEEauthorblockA{\textit{Mechanincal and Mechatronics Engineering} \\
\textit{University of Waterloo}\\
Waterloo, Canada}\\ [0.3cm]
\and
\IEEEauthorblockN{Morteza Babaie, H.R.Tizhoosh}
\IEEEauthorblockA{\textit{Kimia Lab} \\
\textit{University of Waterloo}\\
Waterloo, Canada}
}
\maketitle




\markboth{}
{Shell \MakeLowercase{\textit{et al.}}: Bare Demo of IEEEtran.cls for Computer Society Journals}

\IEEEcompsoctitleabstractindextext{%
\begin{abstract}
Encoded Local Projections (ELP) is a recently introduced dense sampling image descriptor which uses projections in small neighbourhoods to construct a histogram/descriptor for the entire image. ELP has shown to be as accurate as other state-of-the-art features in searching medical images while being time and resource efficient. This paper attempts for the first time to utilize ELP descriptor as primary features for facial recognition and compare the results with LBP histogram on the Labeled Faces in the Wild dataset. We have evaluated descriptors by comparing the chi-squared distance of each image descriptor versus all others as well as training Support Vector Machines (SVM) with each feature vector. In both cases, the results of ELP were better than LBP in the same sub-image configuration.   
\end{abstract}

}

\IEEEdisplaynotcompsoctitleabstractindextext

%
\IEEEpeerreviewmaketitle

\section{Introduction}\label{INT}
\IEEEPARstart{F}{acial} recognition is an important field of computer vision with an ever-growing list of applications including person detection for biometric security, facial autofocus in cameras, and automatic detection and tagging of people in images  \cite{frec, akbari2010legendre}. Extracting distinct and robust face features is considered an essential part of this process. Handcrafted methods, especially local binary patterns (LBP), have proven their discriminative power in face recognition tasks \cite{brahnam2016local}.  
The principal objective of this paper is to examine the application of the Encoded Local Projections (ELP) descriptor  \cite{elppaper} for the facial recognition problem. ELP has demonstrated superior results for image search for three medical datasets. In this paper, the performance of ELP in facial recognition domain is assessed and compared to the LBP as one of the most successful methods for this purpose.

Despite the high accuracy of deep solutions (which require large volumes of data and resources to train), providing alternatives helps broaden the scope of solutions and foster implementations of more generic and applicable feature extraction algorithms. Moreover, in the Wild dataset literature, the top results still belong to the combination of deep network and pure LBP  \cite{xi2016local} or LBP-based methods  \cite{ouamane2015side}, hence, empirically justifying continuous work on handcrafted image descriptors\cite{8285162}.

In this work, to facilitate the comparison of LBP and ELP, the chi-squared distance of image descriptors is calculated and the best match (shortest distance) is reported as a response. 
Further, to compare the learning contribution, a Support Vector Machine (SVM)  classifier is used to classify the LBP and ELP histograms to recognize facial images. 
The results of facial recognition using ELP feature extraction show that ELP histogram has a significant discrimination power for facial recognition and classification which in most cases is superior to LBP results. 

\IEEEpubidadjcol

\section{Background}
From early years of machine vision development, face recognition has gone through numerous phases and efforts regarding its corresponding applicability and complexity  \cite{dolecki2016utility}. Recognition of human faces in real-world applications must be invariant to several changes such as light, poses, expressions, varying resolutions, and occlusions \cite{lu2015learning}. To overcome these challenges, this task has been mainly divided into four sub-divisions: face detection and localization, pre-processing and normalization, feature extraction, and finally feature matching \cite{frec}. Each stage itself can be a challenging problem and several datasets have been created to investigate algorithm development for addressing these challenges \cite{mahalingam2013lbp}.

In this study, we have approached the problem of recognition in a way that is proposed in the Labeled Faces in the Wild dataset: ``Given two pictures, each of which contains a face, decide whether the two people pictured represent the same individual'' \cite{LFWTech}. Holistic and local methods are two main classes of algorithms that have been applied in face recognition literature \cite{lu2015learning}. The most well-known holistic methods, looking at the pixel level across the whole image at one glance, are PCA-Eigenface \cite{turk1991eigenfaces}, LDA  \cite{belhumeur1997eigenfaces} from early stages, and deep solutions in recent years \cite{parkhi2015deep} or SP-RBC for holistic projection-based descriptors \cite{icpram17}. On the other hand, local methods such as LBP \cite{lbp_2}, SIFT \cite{luo2007person}, and HoG \cite{deniz2011face} are applied in small parts (local neighbourhoods) of the image.

Although in light of recent progress in deep learning, it has become extremely difficult to ignore deep networks as a potential solution, the best-reported performance, at least in the Wild dataset, still belongs to the combination of LBP and deep networks \cite{ouamane2015side}. Therefore, it seems that investigation of new methods such as ELP yielding better performance in comparison to LBP could be quite helpful.

\subsection{Local Binary Patterns (LBP)}\label{BG_LBP}
\begin{figure}[!t]
\centering
\includegraphics[width=3.5in]{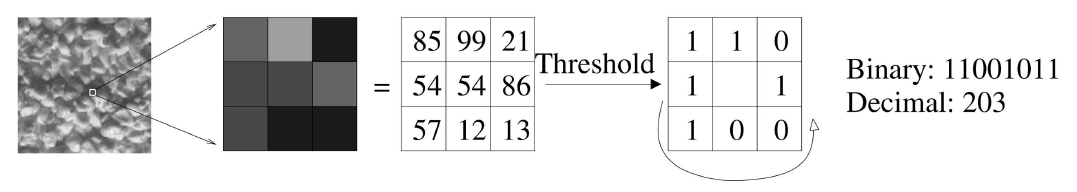}
\caption{The basic LBP operator on a $3\times 3$ window  \cite{lbppaper}}
\label{fig:basic_lbp_operator}
\end{figure}

\begin{figure}[!t]
\centering
\includegraphics[width=3.5in]{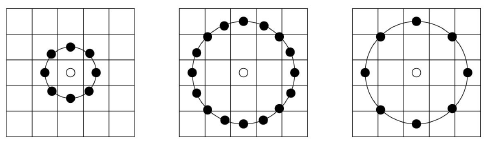}
\caption{The circular windows (8,1), (16,2), and (8,2)   \cite{lbppaper}}
\label{fig:lbp_circular_neighbourhood}
\end{figure}

The LBP operator is among the best texture descriptors with high computational efficiency. Its invariance to grey level changes makes it ideal for image analysis in many domains. A basic operator for a $3\times 3$ neighbourhood is shown in Figure \ref{fig:basic_lbp_operator}. The LBP operator can also use neighbourhoods of different sizes to process textures at different scales. The local neighbourhood, using the notation $(p,r)$, consists of a set of $p$ sampling points evenly spaced on a circle with radius $r$ and centered at the pixel that is to be labelled. When a sampling point lies outside the center of a pixel, bilinear interpolation is used. A visualization of three different  $(p,r)$  is depicted in Figure \ref{fig:lbp_circular_neighbourhood}.  

Every pixel of an image is then assigned a label by the LBP operator by binarizing the neighbourhood of each pixel with the center pixel value as the threshold. The histogram of these labels can then be used as a texture descriptor  \cite{lbppaper}. In 2002, an extension of LBP, \textit{uniform rotationally invariant patterns} was introduced by Ojala et. al \cite{1017623} which not only improves its performance but also reduces the histogram size substantially by merging all uniform patterns. 


\section{Encoded Local Projections}\label{BG_ELP}
Radon transform  \cite{radon1986determination} was introduced in 1986 by J. Radon. This transform defined as  
$R(\theta,\rho)$ along parallel lines $\rho$ and at certain angles $\theta$ can be formulated as
\begin{equation}
\label{radontrans}
R(\rho,\theta)\!=\! \int\displaylimits_{-\infty}^{\infty}\! \int\displaylimits_{-\infty}^{\infty}\! f(x,y)\delta(\rho\!-\!x\cos\!\theta\!-y\!\sin\theta\!)\! \,dx\!\,dy\!
\end{equation}
Here, $\delta(\cdot)$ represents the Dirac delta function.
Every projection $\textbf{p}_\theta$ computed using Radon transform captures a pattern arrangement along $\rho$ at a certain angle $\theta$. Radon transform and its inverse have been widely used in the medical domain \cite{8310144}. In recent years, there has been an increasing interest in applying Radon transform as a descriptor, based on its power to preserve the pattern \cite{8310144}, \cite{tabbone2006new}.

ELP is a histogram-based method that operates on local windows of grey level images  \cite{elppaper}. 
It works by examining and creating binary codes from local projections. 
 To efficiently create a histogram, a relatively small number of projections must be selected in each neighborhood. The projections that capture large changes are the ones that are significant to capture the image details. Flat or homogeneous parts of the image with little to no changes (not sufficiently heterogeneous) are disregarded. Thus, to determine the homogeneity of each window $\textbf{W}_{i,j}$, we may use
\begin{equation}
\label{homogeneity}
H=1-\frac{1}{2^{\mathit{n_{bits}}}}\sqrt{\sum_{\mathit{i}} \sum_{\mathit{j}}(\mathbf{W_{\mathit{ij}}}-m)^{2}},
\end{equation}
\noindent where $m=\textrm{median}_{i,j}(\textbf{W}_{i,j})$ and $n_{bits}$ denotes the number of bits used to encode the image (e.g., $n_{bits} = 8,12,16$).
The projection that lies along the axis of maximum change, is termed the ``anchor projection'' and is denoted by $\theta^*$. The anchor projection ($\theta^*$) is determined using the gradient $\frac{\partial}{\partial \rho} R(\rho_j,\theta_i)$ across parallel lines $\rho_i$:
\begin{equation}
\label{anchorproj}
\theta^* = \underset{i}{argmax}\int_j \frac{\partial}{\partial \rho} R(\rho_j,\theta_i).
\end{equation}
As an efficient implementation, one may simply search for an angle $\theta^*$ whose projection, $\textbf{p}$, has the maximum amplitude  \cite{elppaper}:
\begin{equation}
\label{maxamplitude}
\theta^* = \underset{j}{argmax}[  max[R(\rho_1,\theta_i),...,R(\rho_{|\textbf{p}|},\theta_i)]].
\end{equation}
Since the anchor projection alone may not be enough to encode all the information in an image, additional projections (``adjunct projections'') are computed and anchored to $\textbf{p}_{\theta^*}$ (starting at $\theta^*$): $\Theta=\{\theta^*,\theta^*+\alpha_1,\theta^*+\alpha_2,\theta^*+\alpha_3\}$. These could be equidistant projections: $\Theta=\{\theta^*,\theta^*+\pi/4,\theta^*+\pi/2,\theta^*+3\pi/4\}$. 

These projections need to be encoded in some way to facilitate meaningful and efficient frequency recording. This can be done using ``Min-Max'' encoding \cite{minmaxtizhoosh} applied on the gradient of the projections. For a projection vector $\textbf{p}$ of size $n$ and its derivative $\textbf{p}^{'} = \frac{\partial}{\partial \rho} \textbf{p}$, the binary encoding $\textbf{b}$, $\forall i \in \{1,2,...,n-1\}$, can be given as 
\begin{equation}
\label{binaryencoding}
\mathbf{b}(\mathit{i})=\begin{cases} 1, & \text{if } \mathbf{p}^{'}(\mathit{i}+1)>\mathbf{p}^{'}(\mathit{i})\\ 0, & \text{otherwise} \end{cases}.
\end{equation}

Finally, using the results of all encoded changes, an overall histogram is generated. The ELP histogram represents the key aspects of the images, as each change in the image (edges, corners etc.) corresponds to a change in the local histogram. 

\begin{figure}[!t]
\centering
\includegraphics[width=3.5in]{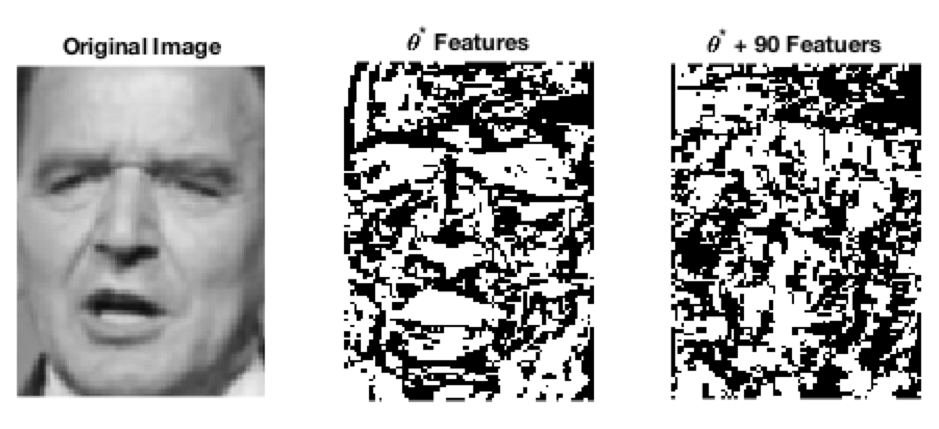}
\caption{Visualization of ELP (from left to right): original image, ELP features for $\theta^{*}$, ELP features for $\theta^{*}+90$.}
\label{fig:elpvisual}
\end{figure}

The ELP histograms can be generated in two different ways: 1) Merged histogram, which counts all binarized derivatives of the projections $\theta$ with $\lvert h \rvert = 2^{\lvert P \rvert}$, and 2) Detached histograms, where all binarized derivatives from each projection are put into a separate histogram and then concatenated into a single longer histogram $h$ with $\lvert h \rvert = \lvert \theta \rvert \times 2^{\lvert P \rvert} $. The detached version of the histogram has slightly higher discrimination power. The ELP descriptor can be displayed by visualizing the decimal values converted from binary patterns as pixel values as shown in Figure \ref{fig:elpvisual}. 
In ELP computations, as is the case for LBP, the entire image can divided into small sub-images, which are encoded independently, and the resulting histograms are concatenated to obtain a final image histogram.

\begin{figure}[!ht]
\centering
\includegraphics[width=\columnwidth]{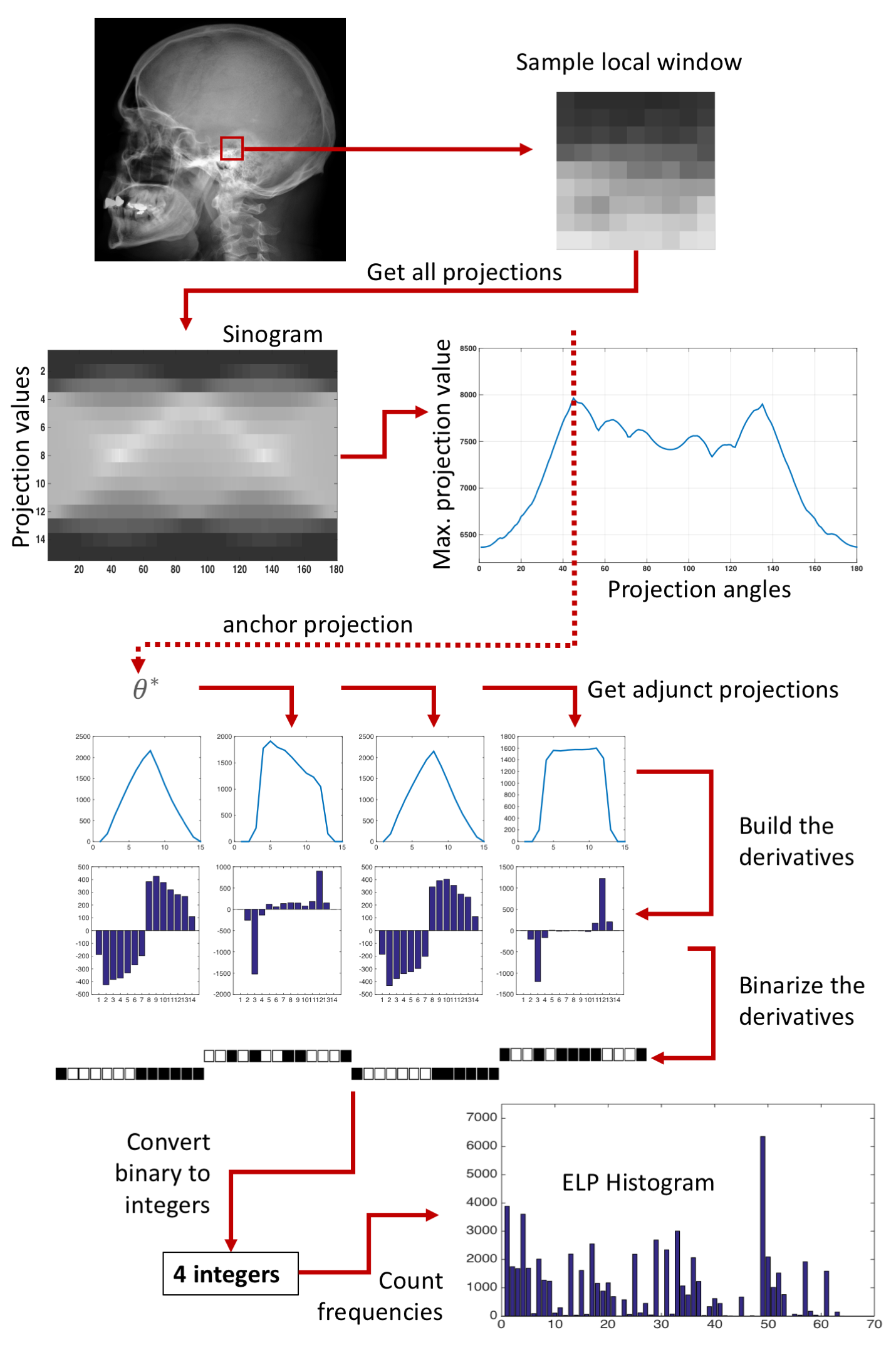}
\caption{Construction of the ELP histogram (based on a modified version of illustration in \cite{elppaper}).}
\label{fig:elpflow}
\end{figure}

Figure \ref{fig:elpflow} shows all steps of the ELP in one window for an x-ray image. 

The idea of using ELP for facial detection is motivated by the fact that ELP captures spatial projection patterns in local windows across an image. Since a face can be regarded as a composition of micro patterns  \cite{lbppaper}, these patterns can be very well captured through spatial projections by the ELP operator.

\section{Experiments}\label{EXP}
The goal of our experiments is to measure the discrimination power of ELP in comparison to LBP for facial recognition. In the first stage, images are encoded using ELP and LBP descriptors, and then standardized to zero mean and unit variance. The encoded version of each image is then compared to every image in the dataset to compute the chi-squared distance. The images with the least chi-squared distance are chosen to be the nearest neighbors and are assumed to have the same class as the input image. 5-Nearest neighbors are evaluated to calculate an accuracy metric. 
After this, a classifier (multi-class SVM) is trained to further determine the discrimination power of the descriptors when applied to the classification task.

\subsection{Dataset}\label{DS}
\begin{figure}[!t]
\centering
\includegraphics[width=2.5in]{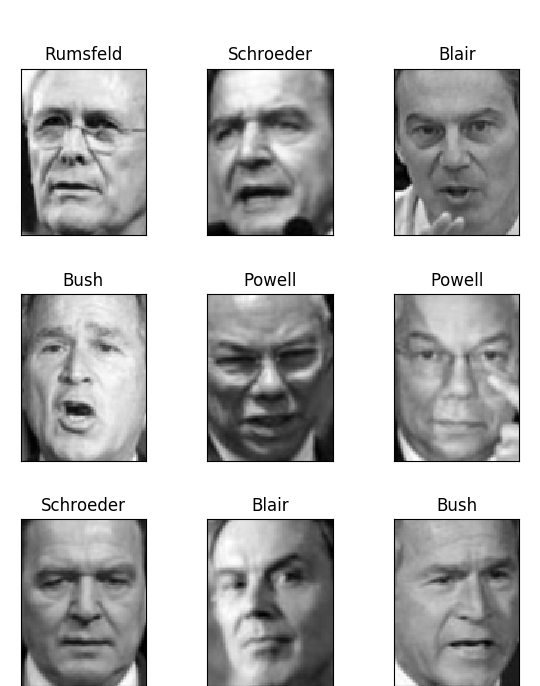}
\caption{Sample images from the LFW dataset}
\label{fig:datasetview}
\end{figure}

To ensure the reproducibility of the results, the publicly available Labelled Faces in the Wild (LFW) dataset is used  \cite{LFWTech}. LFW contains a total of 13233 grayscale images of 5749 different people. About 1680 people have more than 2 images, and only 5 people have more than 100 images in the dataset. Each image is $112 \times 84$ in dimensions.
By considering the fact that the experiments are an initial comparison between ELP and LBP, we have chosen a relatively small subset from LFW by selecting all 5 people with more than 100 samples. As a result, our selected subset contains 1140 images from  5 different identities.  Figure \ref{fig:datasetview} shows sample faces from the dataset. Information about the distribution of the faces can be found in Table 
 \ref{tab:data_dis} .


\begin{table}[!t]
\renewcommand{\arraystretch}{1.3}
\caption{Dataset distribution}
\label{tab:data_dis}
\centering
\begin{tabular}{c||c||c}
\hline
\bfseries Name & \bfseries Number of Images & \bfseries Percentage\\
\hline\hline
Colin Powell      & 236 & 20.7 \\
Donald Rumsfeld    & 121 & 10.6 \\
George W Bush & 530 & 46.5 \\
Gerhard Schroeder & 109 & 9.6 \\
Tony Blair & 144 & 12.6 \\
\hline
\end{tabular}
\end{table}


\subsection{Feature Extraction}\label{FEX}
\begin{figure}[!t]
\centering
\includegraphics[width=3.5in]{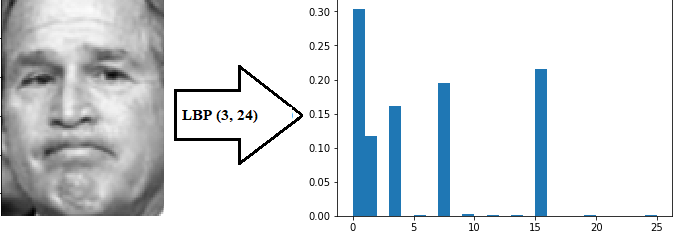}
\caption{Encoding an image into an LBP histogram (radius = 3, sample points = 24)}
\label{fig:lbp_hist}
\end{figure}

\begin{figure}[!t]
\centering
\includegraphics[width=3.5in]{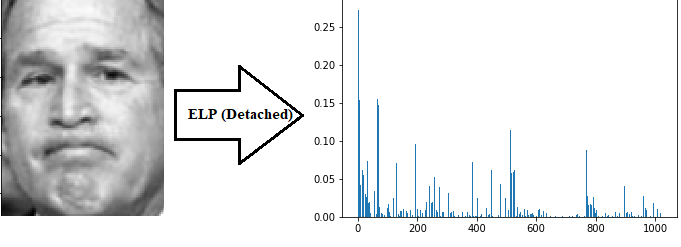}
\caption{Encoding an image into an ELP histogram (window size = 10$\times$10, overlap = 2, homo. threshold = 0.95)}
\label{fig:elp_hist}
\end{figure}

As discussed in Section \ref{EXP}, each image is separately encoded using both ELP and LBP descriptors. For the  rotation invariant and uniform LBP descriptors we have used the Local Binary Patterns class as provided by the scikit-image library in Python. The  major tuning parameters are number of sub-images, radius, and number of points to be evaluated in each neighborhood. Sub-image histograms are then concatenated across the entire image. Figure \ref{fig:lbp_hist} shows the LBP histogram along with the original image. For the ELP descriptor, the major tuning parameters are number of sub-images, window size, and homogeneity threshold. Two separate versions of the descriptor (merged and detached as discussed in section \ref{BG_ELP}) are evaluated. 
Figure \ref{fig:elp_hist} shows the ELP histogram along with the original image.

From Figure \ref{fig:lbp_hist} and Figure \ref{fig:elp_hist}, it can be seen that the lengths of the LBP (26) and ELP detached (1024) are very different. The small length of LBP is due to the use of uniform and rotationally invariant patterns. This concept may also be applicable to ELP, a possibility that future works may consider.

\subsection{Feature Evaluation}\label{FEVAL}
The nearest neighbour search based on chi-squared distance is used to evaluate the relative similarity of image descriptors belonging to the same class. In the first stage, the $k$-NN search approach is used to evaluate ELP and LBP performance. 
 Each target feature is evaluated against each candidate feature to find its five closest neighbors based on the smallest chi-squared distance values. A majority vote is taken among the five nearest neighbors and the resulting feature class is taken to be the output class (based on our experiments 5-NN showed better performance than 1-NN and 3-NN for both LBP and ELP). If the output class is the same class as the target feature, then the search is classified as positive, otherwise negative.

\subsection{Parameter Settings}
Based on \cite{elppaper}, the window size of 10 $\times$ 10 with a unit pixel stride gives the best classification results. More tuning experiments are then run for this experiment with window sizes of 8 to 11, and a stride of 1 pixel. In addition, it is determined that a homogeneity threshold of 0.95 out of tested values \{0.65, 0.75, 0.85, and 0.95\} provides the best results. Both detached and merged histograms produced by ELP descriptor are evaluated. The detached version is found to generally perform better at search and classification. The additional projections $\alpha$ = \{$\theta^*$, $\theta^*$ + 45, $\theta^*$ + 90, $\theta^*$ + 125\} are used following the observations in  \cite{elppaper}. Experiments are run classifying the entire encoded image, the concatenated 2 $\times$ 2 encoded sub-images, and 3 $\times$ 3 sub-images. The ELP results are reported as $\textbf{ELP}_{(w,t)}$ where $w$ is the window size and $t$ is the histogram type ($m$ for merged and $d$ for detached).

For LBP, experiments are run with varying number of sub-images, radius, and the number of sample points. As mentioned in Section \ref{BG_LBP} 'uniform rotationally invariant' LBP is used for all experiments. Numerous different image splitting is used from encoding the whole image to dividing the image into 12 $\times$ 12 sub-images. A smaller radius of 1 pixel with 8 sample points along with a larger radius of 3 and 24 sample points is used. The two sets of \{r, s\} values and different sub-image configurations are tested using the chi-squared distance and classification using SVM. The corresponding results obtained are discussed in Section \ref{SearchResults} and \ref{SVM_Results}. The LBP results are reported as $\textbf{LBP}_{(n,r)}$ where $n$ is the number of neighbors and $r$ is the radius.

\subsection{SVM Classifier}
The C-Support Vector Classifier (SVC) implementation from the \emph{scikit-learn} toolkit is used for SVM classification with an `rbf' kernel  \cite{SVM}. The two main parameters that require tuning for optimization are the penalty parameter $C$ and the kernel coefficient $\gamma$.
A range of $C$ values between $10^{-7}$ to $10^{7}$ and $\gamma$ values between $10^{-5}$ to $10^3$ are experimented with for each iteration to find a suitable value which maximizes classification accuracy. The classifier is trained on 80\% of the overall dataset and tested on the remaining 20\%. The split between the training-testing data is randomly generated.

\section{Results}
The results obtained from the experiments described in Section \ref{EXP} are presented in this section. Table \ref{tab:descriptorLength} reports the descriptor length for different LBP and ELP parameter settings. As can be seen, for both methods,  increasing the number of sub-images leads to higher accuracy. However, since ELP is defined for relatively large windows (around $10 \times 10$), applying on very small sub-images leads to an improper representation in each histogram. By considering the rather small face sizes in the dataset($112 \times 84$), we do not go beyond $3 \times 3$ sub-images ($37 \times 28$). Thus for experimental setting, we start from the whole image and divide it into $2 \times 2$ and then $3 \times 3$ sub-images. In the end, we investigate smaller sub-images (as depicted in Fig. \ref{fig:lbp_sub}) for just LBP, to evaluate its best possible accuracy.   

\subsection{Search}\label{SearchResults}
Search using ELP yields a maximum accuracy of 89.6\% for a window size of $10 \times 10$, pixel stride of 1, homogeneity threshold of 0.95, histogram type of detached, and $3 \times 3$  sub-images. Search using LBP yields a maximum accuracy of 82.5\% for a radius of 3 and 24 sample points at same sub-image configuration ($3 \times 3$) while it catches the best accuracy of 88.5\% at $7 \times 7$  sub-images.

Table \ref{tab:searchresults} summarizes the final accuracy scores obtained for search based on the five nearest neighbor approach using chi-squared distance for histograms when running on an entire image and concatenated histograms of sub-images.

By considering the fact that, LBP accuracy increases with more sub-images (as can be seen in Fig. \ref{fig:lbp_sub}) and peaks at 7 $\times$ 7 sub-images, while ELP cannot be examined in such small windows, still, the retrieval accuracy for ELP is higher than LBP. However, The feature length of the best LBP results (1274 bins for 7 $\times$ 7) is still very small in comparison with the best ELP (9216 bins).

\subsection{Classification}\label{SVM_Results}
Classification using ELP histograms yields a maximum accuracy of 89.9\% for a window size of 10 $\times$ 10, pixel stride of 1, homogeneity threshold of 0.95, histogram type of merged  and $3\times 3$  sub-images. Classification using the LBP histogram yields a maximum accuracy of 79.8\% for a  radius of 3, 24 sample points, and 3 $\times$ 3 sub-images.  

Table \ref{tab:svmresults} summarizes the final accuracy scores obtained using the two descriptors for multi-class classification when running on entire image and concatenated histograms of sub-images. 

\begin{figure}[!t]
\centering
\includegraphics[width=0.99\columnwidth]{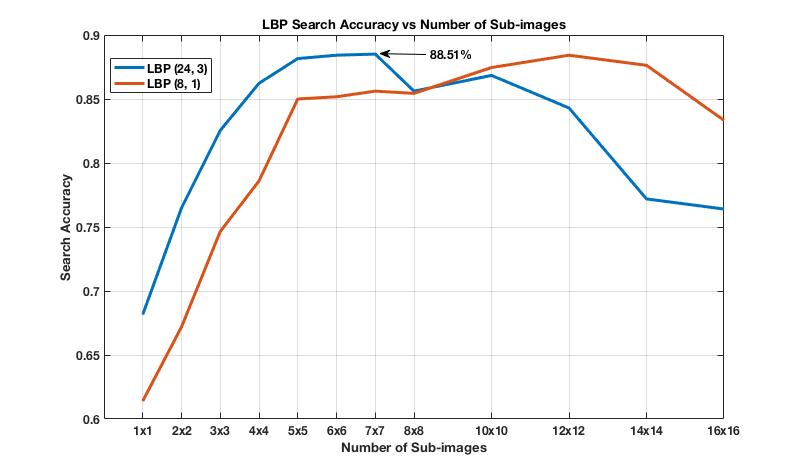}
\caption{Influence of increasing sub-image in LBP}
\label{fig:lbp_sub}
\end{figure}

When the entire image is encoded, classification accuracy for ELP is much higher than LBP. As the number of sub-images increases, both ELP and LBP improve in performance. Maximum classification accuracy of 89\% for LBP is obtained for $6 \times 6$ sub-images. This aligns with the curve seen in figure \ref{fig:lbp_sub} where $6 \times 6$ and $7 \times 7$ sub-images demonstrate high accuracy in search.
%
\begin{table}[!t]
\renewcommand{\arraystretch}{1.3}
\caption{Descriptor length for different parameters}
\label{tab:descriptorLength}
\centering
\begin{tabular}{c||c||c||c||c}
\hline
\textbf{Sub-images} & \textbf{LBP$_{(8,1)}$} & \textbf{LBP$_{(24,3)}$} & \textbf{ELP$_{(10,m)}$} & \textbf{ELP$_{(10,d)}$}\\
\hline\hline
Whole Image  & 10 & 26 & 256 & 1024 \\
2 $\times$ 2    & 40 & 104 & 1024 & 4096 \\
3 $\times$ 3 & 90 & 234 & 2304 & 9216 \\

\hline
\end{tabular}
\end{table}

\begin{table}[!t]
\renewcommand{\arraystretch}{1.3}
\caption{Search accuracy for ELP and LBP}
\label{tab:searchresults}
\centering
\begin{tabular}{c||c||c||c||c}
\hline
\textbf{Sub-images} & \textbf{LBP$_{(8,1)}$} & \textbf{LBP$_{(24,3)}$} & \textbf{ELP$_{(10,m)}$} & \textbf{ELP$_{(10,d)}$}\\
\hline\hline
Whole Image  & 61.4\% & 68.2\% & 71.9\% & 74.3\% \\
2 $\times$ 2    & 67.2\% & 76.5\% & 76.8\% & 82.5\% \\
3 $\times$ 3 & 74.6\% & 82.5\% & 82.1\% & 89.6\% \\
\hline
\end{tabular}
\end{table}

\begin{table}[!t]
\renewcommand{\arraystretch}{1.3}
\caption{Classification accuracy for ELP and LBP}
\label{tab:svmresults}
\centering
\begin{tabular}{c||c||c||c||c}
\hline
\textbf{Sub-images} & \textbf{LBP$_{(8,1)}$} & \textbf{LBP$_{(24,3)}$} & \textbf{ELP$_{(10,m)}$} & \textbf{ELP$_{(10,d)}$}\\
\hline\hline
Whole Image  & 56.1\% & 63.2\% & 73.7\% & 82\% \\
2 $\times$ 2    & 65.4\% & 67.1\% & 79.4\% & 85.5\% \\
3 $\times$ 3 & 67.1\% & 79.8\% & 89.9\% & 89.5\% \\
\hline
\end{tabular}
\end{table}


\section{Discussion and Conclusions}
In this paper, the application of encoded local projections (ELP) has been explored in face recognition domain and its results have been compared with LBP. Once the features (histograms) are extracted, they are used to search for the five nearest neighbors in the dataset based on the chi-squared distance. Both ELP and LBP perform well in this domain. ELP performs better than LBP in general, however, ELP length is much higher than the LBP. The encoded data is then used to train a C-Support Vector Classifier (SVC) for purpose of facial recognition. ELP performs as well as LBP with an accuracy of 89.9 compared to LBP's accuracy of 89. From the results, it can be concluded that ELP is a solid descriptor that can be used for facial recognition as it captures spatial projection patterns in local windows across a face.  

The results obtained in this research provide empirical evidence for the potential of ELP to be used to encode facial images. There are certain experiments that can be done to further evaluate the utility of ELP in facial recognition. It is recommended to tune the three additional projection angles apart from the anchor projection.
In this paper, the projection angles used in  \cite{elppaper}, which apply to medical images, were used, however, different projection angles tuned specifically for facial features might be able to capture more detail. In addition, more projection angles can be added to increase the number of features captured in additional directions. Experimenting with different classifiers can be useful to get better results. Using a neural network can potentially produce better results if the dataset size is large enough.


%



\textbf{Acknowledgment --} The authors would like to thank Shivam Kalra and Aditya Sriram (Kimia Lab; kimia.uwaterloo.ca) for their kind support.

\ifCLASSOPTIONcaptionsoff
  \newpage
\fi



%

\balance
\bibliographystyle{IEEEtran}
\bibliography{reference}

\begin{thebibliography}{10}
\providecommand{\url}[1]{#1}
\csname url@samestyle\endcsname
\providecommand{\newblock}{\relax}
\providecommand{\bibinfo}[2]{#2}
\providecommand{\BIBentrySTDinterwordspacing}{\spaceskip=0pt\relax}
\providecommand{\BIBentryALTinterwordstretchfactor}{4}
\providecommand{\BIBentryALTinterwordspacing}{\spaceskip=\fontdimen2\font plus
\BIBentryALTinterwordstretchfactor\fontdimen3\font minus
  \fontdimen4\font\relax}
\providecommand{\BIBforeignlanguage}[2]{{%
\expandafter\ifx\csname l@#1\endcsname\relax
\typeout{** WARNING: IEEEtran.bst: No hyphenation pattern has been}%
\typeout{** loaded for the language `#1'. Using the pattern for}%
\typeout{** the default language instead.}%
\else
\language=\csname l@#1\endcsname
\fi
#2}}
\providecommand{\BIBdecl}{\relax}
\BIBdecl

\bibitem{frec}
S.~Z. Li and A.~Jain, \emph{Handbook of Face Recognition}, 2nd~ed.\hskip 1em
  plus 0.5em minus 0.4em\relax Springer-Verlag London, 2011.

\bibitem{akbari2010legendre}
R.~Akbari, M.~K. Bahaghighat, and J.~Mohammadi, ``Legendre moments for face
  identification based on single image per person,'' in \emph{International
  Conference on Signal Processing Systems (ICSPS)}, vol.~1, 2010, pp. V1--248.

\bibitem{brahnam2016local}
S.~Brahnam, L.~C. Jain, L.~Nanni, and A.~Lumini, \emph{Local Binary Patterns:
  New Variants and Applications}.\hskip 1em plus 0.5em minus 0.4em\relax
  Springer, 2016.

\bibitem{elppaper}
H.~Tizhoosh and M.~Babaie, ``Representing medical images with encoded local
  projections,'' \emph{IEEE Transactions on Biomedical Engineering}, pp. 1--1,
  2018.

\bibitem{xi2016local}
M.~Xi, L.~Chen, D.~Polajnar, and W.~Tong, ``Local binary pattern network: A
  deep learning approach for face recognition,'' \emph{2016 IEEE international
  conference on Image processing (ICIP)}, pp. 3224--3228, 2016.

\bibitem{ouamane2015side}
A.~Ouamane, M.~Bengherabi, A.~Hadid, and M.~Cheriet, ``Side-information based
  exponential discriminant analysis for face verification in the wild,'' in
  \emph{2015 11th IEEE International Conference and Workshops on Automatic Face
  and Gesture Recognition (FG)}, vol.~02, May 2015, pp. 1--6.

\bibitem{8285162}
M.~D. Kumar, M.~Babaie, S.~Zhu, S.~Kalra, and H.~R. Tizhoosh, ``A comparative
  study of cnn, bovw and lbp for classification of histopathological images,''
  in \emph{2017 IEEE Symposium Series on Computational Intelligence (SSCI)},
  Nov 2017, pp. 1--7.

\bibitem{dolecki2016utility}
M.~Dolecki, P.~Karczmarek, A.~Kiersztyn, and W.~Pedrycz, ``Utility functions as
  aggregation functions in face recognition,'' in \emph{2016 IEEE Symposium
  Series on Computational Intelligence (SSCI)}, Dec 2016, pp. 1--6.

\bibitem{lu2015learning}
J.~Lu, V.~E. Liong, X.~Zhou, and J.~Zhou, ``Learning compact binary face
  descriptor for face recognition,'' \emph{IEEE Transactions on Pattern
  Analysis and Machine Intelligence}, vol.~37, no.~10, pp. 2041--2056, 2015.

\bibitem{mahalingam2013lbp}
G.~Mahalingam and K.~Ricanek, ``Lbp-based periocular recognition on challenging
  face datasets,'' \emph{EURASIP Journal on Image and Video processing}, vol.
  2013, no.~1, p.~36, 2013.

\bibitem{LFWTech}
G.~B. Huang, M.~Mattar, T.~Berg, and E.~Learned-Miller, ``Labeled faces in the
  wild: A database for studying face recognition in unconstrained
  environments,'' University of Massachusetts, Amherst, Tech. Rep. 07-49,
  October 2008.

\bibitem{turk1991eigenfaces}
M.~Turk and A.~Pentland, ``Eigenfaces for recognition,'' \emph{Journal of
  Cognitive Neuroscience}, vol.~3, no.~1, pp. 71--86, 1991.

\bibitem{belhumeur1997eigenfaces}
P.~N. Belhumeur, J.~P. Hespanha, and D.~J. Kriegman, ``Eigenfaces vs.
  fisherfaces: Recognition using class specific linear projection,'' \emph{IEEE
  Transactions on Pattern Analysis and Machine Intelligence}, vol.~19, no.~7,
  pp. 711--720, 1997.

\bibitem{parkhi2015deep}
O.~M. Parkhi, A.~Vedaldi, A.~Zisserman \emph{et~al.}, ``Deep face
  recognition.'' in \emph{BMVC}, vol.~1, no.~3, 2015, p.~6.

\bibitem{icpram17}
M.~Babaie, H.~R. Tizhoosh, S.~Zhu, and M.~E. Shiri, ``Retrieving similar x-ray
  images from big image data using radon barcodes with single projections,'' in
  \emph{Proceedings of the 6th International Conference on Pattern Recognition
  Applications and Methods - Volume 1: ICPRAM,}, INSTICC.\hskip 1em plus 0.5em
  minus 0.4em\relax SciTePress, 2017, pp. 557--566.

\bibitem{lbp_2}
K.~J. Priya and R.~Rajesh, ``A local min-max binary pattern based face
  recognition using single sample per class,'' \emph{International Journal of
  Advanced Science and Technology}, vol.~36, no.~1, pp. 41--50, 2011.

\bibitem{luo2007person}
J.~Luo, Y.~Ma, E.~Takikawa, S.~Lao, M.~Kawade, and B.~Lu, ``Person-specific
  sift features for face recognition,'' in \emph{2007 IEEE International
  Conference on Acoustics, Speech and Signal Processing - ICASSP '07}, vol.~2,
  April 2007, pp. II--593--II--596.

\bibitem{deniz2011face}
O.~D{\'e}niz, G.~Bueno, J.~Salido, and F.~De~la Torre, ``Face recognition using
  histograms of oriented gradients,'' \emph{Pattern Recognition Letters},
  vol.~32, no.~12, pp. 1598--1603, 2011.

\bibitem{lbppaper}
A.~Ahonen, T.~Hadid and M.~Pietikainen, ``Face description with local binary
  patterns: Application to face recognition,'' \emph{IEEE Transactions on
  Pattern Analysis and Machine Intelligence}, vol.~28, no.~12, pp. 2037--2041,
  2006.

\bibitem{1017623}
T.~Ojala, M.~Pietikainen, and T.~Maenpaa, ``Multiresolution gray-scale and
  rotation invariant texture classification with local binary patterns,''
  \emph{IEEE Transactions on Pattern Analysis and Machine Intelligence},
  vol.~24, no.~7, pp. 971--987, July 2002.

\bibitem{radon1986determination}
J.~Radon, ``On the determination of functions from their integral values along
  certain manifolds,'' \emph{IEEE transactions on medical imaging}, vol.~5,
  no.~4, pp. 170--176, 1986.

\bibitem{8310144}
M.~Babaie, H.~R. Tizhoosh, A.~Khatami, and M.~E. Shiri, ``Local radon
  descriptors for image search,'' in \emph{2017 Seventh International
  Conference on Image Processing Theory, Tools and Applications (IPTA)}, Nov
  2017, pp. 1--5.

\bibitem{tabbone2006new}
S.~Tabbone, L.~Wendling, and J.-P. Salmon, ``A new shape descriptor defined on
  the radon transform,'' \emph{Computer Vision and Image Understanding}, vol.
  102, no.~1, pp. 42--51, 2006.

\bibitem{minmaxtizhoosh}
H.~Tizhoosh, S.~Zhu, H.~Lo, V.~Chaudhari, and T.~Mehdi, ``Minmax radon barcodes
  for medical image retrieval,'' \emph{Advances in Visual Computing. IVSC
  2016}, pp. 617--627, 2016.

\bibitem{SVM}
F.~Pedregosa, G.~Varoquaux, A.~Gramfort, V.~Michel, B.~Thirion, O.~Grisel,
  M.~Blondel, P.~Prettenhofer, R.~Weiss, V.~Dubourg, J.~Vanderplas, A.~Passos,
  D.~Cournapeau, M.~Brucher, M.~Perrot, and E.~Duchesnay, ``Scikit-learn:
  Machine learning in {P}ython,'' \emph{Journal of Machine Learning Research},
  vol.~12, pp. 2825--2830, 2011.

\end{thebibliography}

\end{document}